# Associating Natural Language Comment and Source Code Entities

**Sheena Panthaplackel,[1] Milos Gligoric,[2] Raymond J. Mooney,[1] Junyi Jessy Li[3]**
[1]Department of Computer Science
[2]Department of Electrical and Computer Engineering
[3]Department of Linguistics
The University of Texas at Austin
spantha@cs.utexas.edu, gligoric@utexas.edu, mooney@cs.utexas.edu, jessy@austin.utexas.edu

### Abstract

Comments are an integral part of software development; they are natural language descriptions associated with source code elements. Understanding explicit associations can be useful in improving code comprehensibility and maintaining the consistency between code and comments. As an initial step towards this larger goal, we address the task of associating entities in Javadoc comments with elements in Java source code. We propose an approach for automatically extracting supervised data using revision histories of open source projects and present a manually annotated evaluation dataset for this task. We develop a binary classifier and a sequence labeling model by crafting a rich feature set which encompasses various aspects of code, comments, and the relationships between them. Experiments show that our systems outperform several baselines learning from the proposed supervision.

## 1 Introduction

Natural language elements are used to document various aspects of source code. Summaries provide a high-level overview of the functionality of a given code snippet, commit messages describe the code changes that are made between two versions of a software project, and API comments define specific properties of a body of code such as preconditions and the return value. Each of these serves as a crucial mode of communication among developers and is critical to an effective development process. These natural language elements are becoming increasingly prevalent in research within the Natural Language Processing (NLP) community for code summarization (Iyer et al. 2016), commit message generation (Loyola, Marrese-Taylor, and Matsuo 2017), and code generation (Rabinovich, Stern, and Klein 2017; Guu et al. 2017; Yin and Neubig 2017; Yu et al. 2018; Richardson, Berant, and Kuhn 2018; Hayati et al. 2018; Yin and Neubig 2018; Yin and Neubig 2019).

In particular, there is growing interest in cross-modal tasks combining natural language comments and source code. To successfully perform such tasks, it is necessary to understand how elements in a comment relate to elements in the corresponding code. Prior work on detecting inconsistencies between code and comments (Tan et al. 2007; Khamis, Witte, and Rilling 2010; Ratol and Robillard 2017) incorporate task-specific rules to link comment components to aspects of the code. Recent work in automatic comment generation (Hu et al. 2018; Fernandes, Allamanis, and Brockschmidt 2019) rely on an attention mechanism which implicitly approximates the parts of the code that should be attended to when generating certain terms in the comment.

In contrast to these approaches, we formulate a task which aims to learn explicit associations between entities in a comment and elements in the corresponding source code. We believe explicit associations can lead to improved systems for downstream applications. For tasks like code and comment generation, they could serve as a mechanism for supervised attention (Liu et al. 2016) and augment neural models with explicit knowledge, which often leads to significant gains in performance (He et al. 2017; Marcheggiani, Frolov, and Titov 2017; Strubell et al. 2018). Moreover, this provides a means of doing more fine-grained code/comment inconsistency detection, as opposed to the common approach of identifying whether a full comment is inconsistent with a body of code (Tan et al. 2012; Khamis, Witte, and Rilling 2010). Such a system could be a valuable component for automated code/comment maintenance which aims to keep comments consistent with the code they serve to describe. By providing a signal about which elements in the code are referred to by a given entity in the comment, the system can automatically detect when an entity in the comment becomes inconsistent with the code based on changes to these terms in the code.

As an initial step towards learning these associations, we focus on Javadoc `@return` comments, which serve to describe the return type and potential return values that are dependent on various conditions within a given method. We observe that the `@return` comment tends to be more structured than other forms of comments, making it a cleaner data source and consequently, a reasonable starting point for the proposed task. Furthermore, we observe that comments generally describe entities and actions within a given body of code, which map to noun phrases and verb phrases in natural language. In regards to `@return` comments specifically, the return values they characterize are typically entities and conditions related to entities (e.g., input parameters, program state). Because we focus on such comments in this paper, we target noun phrase entities within `@return` com-

```
/* @return Snapshot or null when there are problems
    reading it. */
public ConfigRepo.Snapshot getLatestConfig() {
    if (latestConfig == null) {
        try {
            updateConfigSnapshot();
        } catch ( InterruptedException  e ) {
            Thread.currentThread().interrupt();
        }
    }
    return latestConfig;
}
```

Figure 1: Example from the node-sharing-plugin project with the boxed/bolded tokens in the code being associated with the underlined NP in the comment.

```
/* @return the opcode of the current bytecode */
public int next() {
    final  int   opcode  =  currentBC ();
    setBCI(_nextBCI);
    return  opcode ;
}
```

Figure 2: Example from adriaanm-maxine-mirror project with the boxed/bolded tokens in the code being associated with the underlined NP in the comment.

ments as a first step. Illustrated in Figures 1 and 2, given a noun phrase in a comment (underlined), the task is to identify code tokens with which they are associated (bolded).

Learning to automatically resolve associations between comments and code, however, is challenging in terms of data collection. Acquiring annotated data for code/language tasks is difficult since it would require expertise in comprehending source code in a particular programming language. Moreover, it is challenging to collect a high-quality parallel corpus containing source code and natural language because the data in large online code bases is inherently noisy (Yao et al. 2018; Yin et al. 2018). In this paper, we propose a novel approach, requiring no human annotation, for obtaining noisy supervision for this task from GitHub, using the platform's commit history feature. We show that this noisy supervision provides a valuable training signal.

To establish ground work for future research on this task, we design a set of highly salient features with relatively simple models. We propose two models which are trained on noisy data and evaluated on a manually labeled test set.[1] The first is a binary classifier which independently makes a classification for each element in a given code snippet on whether or not it is associated with a specified noun phrase in the corresponding comment. Our second model is a sequence labelling model, specifically a conditional random field (CRF) model, which jointly assigns labels to elements in the code, where the labels denote whether or not an element is associated with the specified noun phrase. We design a set of novel features capturing contextual representations, cosine similarity, and the API and grammar relevant to the programming language.

Trained on noisy data, the two models outperform baselines by wide margins, with the binary classifier attaining an F1 score of 0.677 and the CRF attaining an F1 score of 0.618, achieving 39.6% and 27.4% improvement from baselines, respectively. We demonstrate the value of noisy supervision by showing improved performances of our models as the size of noisy training data increases. Additionally, through an ablation study, we highlight the utility of the features that are consumed by our models. The main contributions of this paper are summarized as follows:

- The new task of associating entities in natural language comments with elements in source source code, with a manually labeled evaluation dataset for this task;
- A technique for obtaining noisy supervision from histories of software changes and machine learning systems that leverage this form of supervision;
- A novel feature set that captures characteristics of code and comments as well as the relationships between them, used in models that can serve as baselines for future work.

## 2 Task

Given a noun phrase (NP) in a comment, the task is to classify the relationship between the NP and each candidate code token in the corresponding code as either associated or not associated. The candidates include all tokens other than Java keywords (e.g., `try`, `public`, `throw`), operators (e.g., `=`), and symbols (e.g., brackets, parentheses); these elements are related to the programming language syntax and are commonly not described in comments. For instance, in Figure 2, the tokens `int`, `opcode`, and `currentBC` are associated with the NP "the current bytecode" but `int` (the return type), `setBCI`, and `_nextBCI` are not.

This task shares similarities with anaphora resolution in natural language texts, including ones that explicitly refer to antecedents (coreference) as well as ones linked by associative relations (bridging anaphora) (Mitkov 1999). In such a setting, the selected noun phrase within the comment is the anaphor, and tokens belonging to the source code serve as candidate antecedents. However our task is distinct from either in that it requires reasoning with respect to two different modalities (Allamanis et al. 2015; Loyola, Marrese-Taylor, and Matsuo 2017; Allamanis et al. 2018). In Figure 1, "problems" explicitly refers to `e`, but we need to know that `InterruptedException` is its type, which is a kind of `Exception`, and that `Exception` is a programming term for "problems." Further, in our setting, an NP in the comment could be associated with multiple, distinct elements in the source code that do not belong to the same co-reference "chain." For these reasons, we frame our task broadly as *associating* a noun phrase in a natural language comment with individual code tokens in the corresponding body of code.

We use the Java programming language and Javadoc comments, namely, `@return` comments, in this work; however

---

[1] The full dataset (including the annotated test set) and implementation are available at https://github.com/panthap2/AssociatingNLCommentCodeEntities.

this task and the methodology can be extended to other programming languages. For instance, Python Docstring and C# XML documentation comments serve similar purposes.

## 3 Data Source

We use one of the most well-structured types of comments for Java, namely, comments tagged with `@return` that are part of the Javadoc documentation (figures 1 and 2 are two examples of such comments).

The content in the `@return` tag[2] provides a fairly comprehensive overview of the functionality of the corresponding method, since these comments describe the output, which is computed by the various statements that make up the method. In contrast, content in other Javadoc tags are generally more narrow in scope, and unstructured comments tend to be long and high-level in nature, making it difficult to map directly to elements in code (see supplementary material for examples). We leave it to future work to extend the proposed task to other types of comments. Hereon, when we refer to comments, we are referring to content attached to `@return` tags.

We construct a dataset by extracting examples from all commits of popular open-source projects on GitHub. We rank the projects by the number of stars, and used the top ∼1,000 projects, as they are considered to be of higher quality (Jarczyk et al. 2014). Each example we extract consists of a code change to a method body as well as a change to the corresponding `@return` comment.

## 4 Noisy Supervision

The core idea of our noisy supervision extraction method is to utilize revision histories from software version control systems (e.g., Git), based on prior research showing that source code and comments co-evolve (Fluri, Würsch, and Gall 2007). Essentially, entities in comments have a higher chance of being associated with entities in source code if they were edited "at the same time", which can be approximated by "at the same commit". Therefore, mining such co-edits allow us to obtain noisy supervision for this task: we use the version control system Git to isolate parts of the code and comment that are added and deleted together.

### 4.1 Supervision setup

**Additions** Based on the intuition that the parts of the code that are added are likely to be associated with the parts of the comment that are also added, we assign noisy labels to the code tokens. Namely, we label any code token that is added in a given commit as associated with the NP that is introduced in the comment within the same commit, and we label all other code tokens as not associated with the NP. These positive labels are noisy since a developer may also make other code changes that are not necessarily relevant to the NP that is added. On the other hand, the negative labels (not associated) have minimal noise, since code tokens that are retained from the previous version of the code are unlikely to be associated with an NP that does not exist in the previous version of the comment. This set of examples we collect from additions constitute our *primary dataset*.

[2] https://docs.oracle.com/javase/8/docs/technotes/tools/windows/javadoc.html

Figure 3: Diff from a commit of the adriaanm-maxine-mirror project. Green lines starting with '+': added; red lines starting with '-': removed. Two examples can be extracted from this, one for the deleted case (i.e., parts of the comment+code present only in the previous version) and one for the added case (i.e., parts of the comment+code present only in the new version). Based on the supervision provided by the diff, the bolded code tokens are automatically labeled as associated with the underlined NP in the comments.

**Deletions** Theoretically, if we assume that the code tokens that are deleted are associated with an NP that is deleted from the comment, we may be able to extract one more example from each commit. However, deleted NPs are much more subtle in this respect than added NPs. As stated above, since the added NP does not exist in the previous version, it is unlikely that code tokens that existed previously are associated with it. On the other hand, since the deleted NP does exist in the previous version, we cannot reliably claim that a token in the code that is unchanged between versions is not associated with the NP. This could consequently lead to more noise for the negative label in addition to the noise that inherently exists for the positive label. For instance, in the deleted example from Figure 3, _nextBCI is automatically labeled as not associated with the deleted NP "the next bytecode" even though it is arguably associated. Hence we separate such examples from our primary dataset and form another set of examples we refer to as *the deletions dataset*.

### 4.2 Processing

We examine the two versions of the code and comment in a commit: *before* commit and *after* commit. Using spaCy,

we extract NPs from the two versions of the comment, and using the javalang library, we tokenize the two versions of the code. Using the difflib library, we compute the *diff* between the NPs in the two versions of the comment as well as the diff between the two versions of the tokenized code sequences. These *diff*s are marked with plus and minus signs for each changed line, as shown in Figure 3.

From the diffs, we identify the NPs and code tokens that are unique to either the *before* or the *after* versions of the comment and code respectively, allowing us to construct two pairs in the form *(NPs, associated code tokens)*. If either the extracted NPs or list of associated code tokens is empty, we discard the pair. Additionally, we discard pairs consisting of more than one NP to obtain unambiguous training data for determining which code tokens should be associated with which NP. Therefore, the final set of pairs are in the form *(NP, associated code tokens)*. Note that for any token in the associated code tokens, if it is not a common Java type (e.g., `int`, `String`), we also treat any other token in the sequence with the same literal string as associated.

We then go back to the *before* and *after* versions of the code (excluding Java keywords, operators and symbols, c.f. Section 2). We tokenize the code sequence and label any token that is not present in the associated code tokens as not associated. Following this procedure, each example consists of an NP and a sequence of labelled code tokens. The example extracted from the previous version (*before*) is added to the deletions dataset and the one from the new version (*after*) is added to the primary (additions) dataset.

### 4.3 Filtering

While large code bases such as GitHub and StackOverflow offer vast amounts of data, it is challenging to obtain large quantities of high-quality parallel data for tasks involving source code and natural language for a number of reasons such as significant levels of noise (Yin et al. 2018) and code duplication (Allamanis 2019). Prior work has addressed this problem by filtering out low-quality examples with classifiers trained on manually labeled data (Iyer et al. 2016; Yao et al. 2018; Yin et al. 2018). However, since acquiring manually labeled data is difficult for this task, we choose to apply heuristics, as done in prior work (Allamanis, Peng, and Sutton 2016; Hu et al. 2018; Fernandes, Allamanis, and Brockschmidt 2019). We impose constraints to filter out noisy examples, including duplicates, trivial cases, and examples consisting of unrelated code and comment changes.

We define trivial cases as those examples involving single-line methods which consist of only a few code tokens that are all likely to be associated with the NP as well as those examples in which all associations can be resolved with a simple string matching tool.

Additionally, after manually inspecting a sample of approximately 200 examples, we establish heuristics to minimize the number of examples with unrelated code and comment changes: (1) those that have lengthy methods or a substantial number of code changes which are likely not to all be correlated with the comment; (2) cases with changes to the code and comment that are related to re-formatting, typo fixes, and simple rephrasing; (3) examples involving com-ment changes entailing verb phrases as the corresponding code changes could be related to these phrases rather than the NP. In addition, since we focus on the `@return` tag that describes the return value of a Java method, we eliminate examples with code changes that do not include either a change to the return type or at least one return statement. See supplementary material for specific parameters and the number of examples discarded by each heuristic.

| | | Candidate Code Tokens | | |
|---|---|---|---|---|
| **Partition** | **Examples** | **Total** | **Unique** | **Average** |
| Train | 776 | 23,188 | 5,908 | 29.9 |
| Validation | 77 | 2,488 | 911 | 32.3 |
| Test (annotated) | 117 | 3,592 | 1,266 | 30.7 |
| Deletions | 867 | 25,203 | 6,186 | 29.1 |

Table 1: Number of examples, total and unique candidate tokens, and average number of candidate tokens per example, for each partition of the dataset.

Applying such heuristics substantially reduced the size of our dataset. However, we determined such filtering to be necessary after manually inspecting 200 examples and observing significant noise, and finding that is consistent with aforementioned prior work, which pointed out that the levels of noise in large code bases are too substantial to learn from without aggressive filtering and pre-processing.

Upon filtering, we partition our primary dataset into train, test, and validation sets, shown in Table 1. Based on the training set, the median number of words in the NP is 2 with an interquartile range (IQR, difference between 25% and 75% percentile)[3] of 1, the median number of code tokens is 25 with IQR 21, and the median number of associated code tokens is 10 with IQR 13.

### 4.4 Test Set

The 117 examples in the test set were annotated by one of the authors who has 7 years of experience with Java. During pilot studies, two annotators jointly examined a sample set of method/comment pairs before converging on the criteria that were used for annotation. The standards used to identify a code token as associated include: whether it is directly referred to by the NP; it is an attribute, type, or method corresponding to the entity referred to by the NP; it is set equal to the entity referred to by the NP; and if an update to the NP would be required if the token is changed. See supplementary material for examples of annotations. To assess the quality of the annotations, we asked a graduate student, who is not one of the authors and has 5 years of Java experience, to annotate 286 code tokens (from 25 examples in the test set) that are labeled associated under the noisy supervision. The Cohen's kappa score between the two sets of annotations is 0.713, indicating satisfactory agreement.

## 5 Representations and features

We design a set of features that encompasses surface features, word representations, code token representations, cosine similarity between terms, code structure, and the Java

---
[3]We report IQR since the distributions are not normal.

API. Our models leverage the 1,852-dimensional feature vector that results from concatenating these features.

**Surface features.** We incorporate two binary features, subtoken matching and presence in return statement, which we also use in two of the baseline models that are discussed in the next section. The subtoken matching feature indicates that a candidate code token matches exactly with a component of the given noun phrase, at the token-level or subtoken-level (ignoring case). Subtokenization refers to splitting camel case that is commonly used in Java (e.g., `maxResult` would be split into `max` and `result`). The presence in return line feature indicates whether a candidate code token appears in a return statement or matches exactly with any token that appears in a return statement.

**Word and code token representations.** In order to derive representations of terms in the comment and code, we pre-train character-level and word-level embeddings for the comment and character-level, subtoken-level, and token-level embeddings for the code. These 128-dimensional embeddings are trained on a much larger corpus, consisting of 128,168 `@return` tag/Java method pairs that are extracted from GitHub. The pre-training task is to generate `@return` comments for Java methods using a single-layer, unidirectional SEQ2SEQ model (Sutskever, Vinyals, and Le 2014). We use averaged embeddings to derive representations for the NP and candidate code token. Additionally, in order to provide a meaningful context, we average the embeddings corresponding to the full `@return` comment as well as the embeddings corresponding to the tokens in the same line in which the candidate token appears.

**Cosine similarity.** Recent work has used joint vector spaces for code/natural language description pairs and has shown that a body of code and its corresponding description have similar vectors (Gu, Zhang, and Kim 2018). Since the content of `@return` comments often mention entities in the code, rather than modeling a joint vector space, we project the NP into the same vector space of the code by computing its vector representations with respect to the embeddings trained on Java code. We then compute the cosine similarity between the NP and the candidate code token at the token-level, subtoken-level, and character-level. The same procedure is followed to compute the cosine similarity between the NP and the line in the code on which the candidate code token appears.

**Code structure.** An abstract syntax tree (AST) captures the syntactic structure of a given body of code in tree form, as defined by Java's grammar. Using javalang's AST parser, we derive the AST corresponding to the method. In order to represent properties of the candidate code token with respect to the overall structure of the method, we extract the node types of its parent and grandparent and represent them with one-hot encodings. This provides deeper insight into the role of a candidate code token within the broader context of the method by conveying details such as whether it appears within a method invocation, a variable declaration, a loop, an argument, a try/catch block, and so on.

**Java API.** We use one-hot encodings to represent features related to common Java types and the `java.util` package, which is a collection of utility classes, such as `List`, that we found to be used frequently. We hypothesize that these features could shed light into patterns that are exhibited by these frequently occurring tokens. To capture local context, we also include Java-related characteristics of code tokens adjacent to the candidate token such as whether it is a common Java type or one of the Java keywords.

## 6 Models

We develop two models representing different ways to tackle our proposed task: binary classification and sequence labeling. We also formulate multiple rule-based baselines.

### 6.1 Binary Classification

Given a sequence of code tokens and an NP in the comment, we independently classify each token as associated or not associated. Our classifier is a feedforward neural network with 4 fully-connected layers and a final output layer.[4] As input, the network accepts a feature vector corresponding to the candidate code token (discussed in the previous section) and the model outputs a binary prediction for that token.

### 6.2 Sequence Labeling

Given a sequence of code tokens and an NP in the comment, we jointly classify the tokens regarding whether or not they are associated with the NP. The intuition behind structuring the problem this way is that the classification of a given code token can often depend on classifications of nearby tokens. For instance, in Figure 3c, the `int` token that denotes the return type of the `next()` function is not associated with the specified NP, whereas the `int` token that is adjacent to `opcode` is considered to be associated because `opcode` is associated, and `int` is its type.

In order to re-establish the consecutive ordering of the original sequence, we inject removed Java keywords and symbols back into the sequence and introduce a third class which serves as the gold label for these inserted tokens. Specifically, we predict the three labels: *associated*, *not associated*, and a pseudo-label *Java*. Note that we disregard the classifications of these tokens during evaluation, i.e., if this pseudo-label is predicted for any other code token at test time, we automatically assign it to be not associated (on average, this happens ∼1% of the time). We construct a CRF model (Lample et al. 2016) by applying a neural CRF layer on top of a feedforward neural network that resembles that of the binary classifier in structure, except that the network accepts a matrix consisting of the feature vectors of all the tokens in the method.[5]

### 6.3 Model Parameters

The 4 fully-connected layers have 512, 384, 256, and 128 units. Dropout is applied to each of these with probability 0.2. We terminate training if there is no improvement in the F1 score on the validation set for 5 consecutive epochs (after

---

[4] We experimented with a logistic regression model as a classifier; however, it did not perform as well as the neural network.

[5] We experimented with a non-neural CRF model using sklearn-crfsuite; however, it did not perform as well as the neural model.

10 epochs), and we use the model corresponding to the highest validation F1 score up till that point. We implemented both models with TensorFlow.

### 6.4 Baselines

**Random.** Random classification of a code token as associated or not based on a uniform distribution.

**Weighted random.** Random classification of a code token as associated or not associated based on the probabilities of the associated and not associated classes as observed from the training set which are 42.8% and 57.2% respectively.

**Subtoken matching.** Any token for which the *subtoken matching* surface feature (introduced in the previous section) is set to be true is classified as associated while all other tokens are classified as not associated. Note that there will never be a case in which *all* associated code tokens will match at the token-level or subtoken-level with the noun phrase. We removed such trivial examples from the dataset during filtering because they can be resolved with simple string-matching tools and are not the focus of this work.

**Presence in return statement.** Any token for which the *presence in a return statement* surface feature (discussed in the previous section) is set to be true is classified as associated and all other tokens are classified as not associated.

## 7 Results

We evaluate our models using micro-level precision, recall, and F1 metrics. That is, we evaluate our models at the token-level, on the 3,592 NP-code token pairs in the test set. All reported scores are averaged across three runs. In the following sections, we discuss results from training on just the primary training set, results from incorporating the deletions dataset into training, and results from an ablation study of the features used by the binary classifier and CRF model.

### 7.1 Training on Primary Dataset

The results of the three baselines and our models are given in Table 2. Our analysis is primarily based on the results on the annotated test set, and we show the results from the unannotated set simply for completeness. Relative to scores from the unannotated set, the models tend to achieve lower precision scores and higher recall scores with the annotated set. This is expected since the number of tokens with the gold label *associated* was reduced during the annotation procedure.

Both of our models outperform the baselines by wide margins. See supplementary material for sample output from the binary classifier. Although the recall score of the CRF is slightly higher than that of the binary classifier, it is clear that the binary classifier performs better overall with respect to the F1 score. This may be due to the fact that the CRF requires additional parameters to model dependencies which may not be set accurately, given the limited amount of example-level data in our experimental setup. Furthermore, while we expect the CRF to be more context-sensitive than the binary classifier, we do incorporate many contextual features (embeddings of surrounding and neighboring tokens, similarity of context with the NP, and Java API knowledge of neighboring tokens) with the binary classifier. With error analysis we found that the CRF model tends to make mistakes over tokens following Java keywords, as well as tokens that appear later in a method. This indicates that the CRF model could be struggling to reason over longer range dependencies and over longer sequences. Additionally, in contrast to the binary classification setting, Java keywords are present in the sequence labeling setting, so the CRF model must reason about many more code tokens than the binary classifier.

| Testset | Model | Prec. | Recall | F1 |
|---|---|---|---|---|
| Annotated | Random | 0.321 | 0.472 | 0.382 |
| | Weighted random | 0.338 | 0.428 | 0.378 |
| | Subtoken matching | 0.567 | 0.338 | 0.428 |
| | Presence in return line | 0.515 | 0.458 | 0.485 |
| | Binary Classifier | **0.574** | 0.654 | **0.610** |
| | CRF | 0.484 | **0.663** | 0.559 |
| Unannotated | Random | 0.396 | 0.498 | 0.441 |
| | Weighted random | 0.395 | 0.425 | 0.409 |
| | Subtoken matching | 0.583 | 0.294 | 0.391 |
| | Presence in return line | 0.561 | 0.423 | 0.482 |
| | Binary Classifier | **0.647** | 0.633 | **0.640** |
| | CRF | 0.521 | 0.581 | 0.533 |

Table 2: Micro precision, recall, and F1 scores after training on the primary training set, evaluated on the annotated and unannotated test sets. The differences between F1 scores within the same test set are statistically significant based on a signed rank t-test, with $p < 0.01$.

### 7.2 Augmenting Training with Deletions

We increase the training set by adding data in stages from the deletions dataset. The results from training the binary classifier and CRF on these new supplemented datasets are shown in Table 3. For the binary classifier, adding 500 and 867 deleted examples seems to provide a significant boost in F1, and for the CRF model, adding any amount of deleted examples leads to improved performance. This indicates that our models can learn from data that we consider to be more noisy than the primary training set that we collect. Since we are able to find value in both the added case as well as the deleted case corresponding to a given commit, we are able to substantially increase the upper bound on the amount of data that can be collected to train models that perform our proposed task. This is particularly encouraging given how difficult it is to obtain a large amount of high-quality data for this task. Despite having extracted examples from methods in source code files across all commits of more than 1,000 projects, we only acquire a total of 970 examples from added cases after filtering for noise. By including the 867 examples from deleted cases, we increase this number to 1,837. While this is still a relatively small number, we expect the potential size to increase substantially as the scope of the task is extended to other comments beyond the `@return` comments that we focus on in this paper for an initial study.

### 7.3 Ablation Study

We conduct an ablation study on the binary classifier trained on the primary dataset in order to analyze the impact of the features we introduce. We ablate cosine similarity, embed-ding, and the Java-related features. The embedding features include code embeddings (i.e., the embeddings corresponding to the candidate code token and the tokens in the line of the method) and comment embeddings (i.e., the embeddings corresponding to the NP and `@return` comment). [6] Based on the results shown in Table 4, all of these features contribute in a positive manner towards the performance of the full model, with respect to the F1 metric.

| Model | # of Deleted Examples | Precision | Recall | F1 |
|---|---|---|---|---|
| Binary | 0 | 0.574 | 0.654 | 0.610 |
| | 100 | 0.572 | 0.639 | 0.603 |
| | 200 | 0.554 | 0.689 | 0.614 |
| | 500 | 0.624 | 0.693 | 0.655 |
| | 867 | **0.644** | **0.715** | **0.677** |
| CRF | 0 | 0.484 | 0.663 | 0.559 |
| | 100 | 0.482 | 0.736 | 0.582 |
| | 200 | 0.512 | 0.685 | 0.585 |
| | 500 | 0.504 | 0.740 | 0.599 |
| | 867 | **0.528** | **0.745** | **0.618** |

Table 3: Micro precision, recall, and F1 scores after training on the primary training set and a varying number of deleted examples, tested on the annotated test set.

| Model | Precision | Recall | F1 |
|---|---|---|---|
| Full | 0.574 | 0.654 | **0.610** |
| - code embeddings | 0.519 | 0.617 | 0.562 |
| - comment embeddings | 0.523 | **0.675** | 0.587 |
| - cosine similarity | **0.582** | 0.613 | 0.597 |
| - Java API & AST | 0.543 | 0.641 | 0.588 |

Table 4: Micro precision, recall, and F1 scores for the binary classifier upon ablating certain features, tested on the annotated test set. All differences in F1 are statistically significant based on a signed rank t-test, with p < 0.01.

## 8 Related Work

Prior work examines a task involving grounding noun phrases within a dialogue system to a programming environment (Li and Boyer 2015; Li and Boyer 2016). The noun phrases are extracted from interactions between students and tutors, and the programming environment hosts students' code. The nature of their work resembles coreference resolution as the goal is to identify the entities within the programming environment that are referred to by a given noun phrase. Within the dialogue, students and tutors discuss implementation details that pertain to specific entities in the code, which makes coreference resolution an appropriate way to frame the task. In contrast, the subject of their work—comments that accompany source code—often describe high-level functionality rather than implementation details. Since multiple components in the code interact to compose the functionality, there could be entities in the code that are directly or indirectly referred to by a given element in the comment. Because their data and implementation were not publicly available, we could not do any further comparisons between the tasks and approaches.

---

[6] Models without embeddings also do not include cosine similarity, as the latter depends on the embeddings.

Fluri, Würsch, and Gall (2007) studies a variant task related to mapping a single source code component (e.g., class, method, statement) to a line or block comment based on distance metrics and other simple heuristics. In contrast, our approach treats both code and comments at a more fine-grained level—we model code at the token level and consider NPs in comments. Additionally, under our framework, multiple code tokens can be mapped to the same NP, and these mappings are learned from data extracted from changes.

Liu et al. (2018) introduces a task related to linking different change intents contained in a single commit message to source code files in a software project which have changed within the commit. While this entails associating components within a natural language message to source code much like the task we propose, the associations we are interested in occupy a higher level of granularity. Namely, we focus on NPs in comments while their work is concerned with sentences and clauses in commit messages, and the unit of classification is per individual code token in our case while it is per file in their work. Additionally, the dataset that is constructed in their work extracts source code files that have changed and commit messages, which are newly written for each commit. On the contrary, for our task, we collect examples that encompass changes in both source code and comments, which co-evolve.

The procedure we follow to extract examples from Git version control system is similar to the approach taken by Faruqui et al. (2018) to build a corpus of Wikipedia edits based on Wikipedia's edit history. They extract samples from insertions of contiguous text to sentences in Wikipedia articles, and these examples are expected to demonstrate how natural language text is typically edited. In contrast, we do not limit to just insertions or require edits to be contiguous. Furthermore, we strive to collect examples that demonstrate how two modalities are edited together.

## 9 Conclusion

In this paper, we formulated the task of associating entities in Javadoc comments with elements in Java source code. We proposed a novel approach for obtaining noisy supervision for this task, and we presented a rich set of features that aim to capture aspects of the code, comments, and the relations that hold between them. Based on evaluation conducted on a manually labeled test set, we showed that two different models trained on such noisy data can significantly outperform multiple baselines. Moreover, we demonstrated the potential for learning from noisy data by showing how increasing the size of the noisy training data can lead to improved performance. We also highlighted the value of our feature set through an ablation study.

## Acknowledgments

We thank Pengyu Nie, Angela Lin, Julia Strout, Jialin Wu, Aishwarya Padmakumar, Prasoon Goyal, and reviewers for their feedback on this work. This work was partially supported by a Google Faculty Research Award and the NSF Grant Nos. CCF-1652517 and IIS-1850153.

```
/**
 * Check the beanFactory to see whether the bean
 * named beanName already exists. Accounts for
 * the fact that the requested bean may be "in
 * creation", i.e.: we're in the middle of servicing
 * the initial request for this bean. From JavaConfig's
 * perspective, this means that the bean does not
 * actually yet exist, and that it is now our job to
 * create it for the first time by executing the logic
 * in the corresponding Bean method. Said another
 * way, this check repurposes
 * ConfigurableBeanFactory#isCurrentlyInCreation
 * to determine whether the container is calling this
 * method or the user is calling this method.
 * @param beanName name of bean to check for
 * @return true if beanName already exists in
 * the factory
 */
boolean factoryContainsBean(String beanName) {
    return beanFactory.containsBean(beanName)
        && !beanFactory.isCurrentlyInCreation(beanName);
}
```

Figure 4: Example from spring-framework project comparing the nature of unstructured free text, content attached to `@return`, and content attached to `@param`.

## A  Comparison Between Parts of a Javadoc Comment

Figure 4 displays examples of text in unstructured free comment text, in the `@return` tag, and in the `@param` tag. The unstructured text spanning lines 2-14 is extremely long and consists of many NPs which cannot be mapped to tokens in the method. The text attached to the `@param` tag focuses on the input to the method rather than the purpose the method serves. Meanwhile, the text attached to the `@return` tag captures the essence of the method while remaining concise.

## B  Filtering

The number of examples filtered out from the primary dataset by each heuristic is shown in Table 5. In this section, we discuss specific parameters used for many of these filtering cases.

**Poorly maintained projects** We extract the majority of examples from the top 1,000 projects in order to minimize the use of poorly maintained projects that may have inconsistent code and comments as a result of developers not updating comments when making code changes (Jiang and Hassan 2006).

**Trivial cases** We do not consider methods with less than 4 lines of code as we observe that such methods generally have only one line in the method body, and we believe it would be trivial to classify the few code tokens present in such a method. Additionally, we remove cases in which the lexical string of *all* the associated code tokens match some component of the given NP in the comment, either at the token-level or *subtoken*-level (e.g., for NP "max result" and code token `maxResult`).

| Category | Filter | # Discarded |
|---|---|---|
| Trivial | Short methods | 5,218 |
| | Lexical string matching | 828 |
| Unrelated | No return statement/type change | 3,709 |
| | Long methods | 692 |
| | Many added code tokens | 67 |
| | Many diff lines | 6 |
| | Added VP | 397 |
| | Re-formatting/typos/rephrasing | 1,275 |
| Other | Duplicates | 546 |
| | Multiple NPs added | 1,574 |
| | No NPs added | 856 |
| | No code tokens added | 167 |
| | | 15,335 |

Table 5: Number of examples filtered out of the primary dataset by each heuristic. Prior to filtering, there are 16,305 examples, and following filtering, there are 970 examples.

**Unrelated code and comment changes** Because we are focusing on the `@return` tag of the Javadoc comment, code changes involving a return statement or return type are more likely to be relevant to the change in the comment, and so we only consider examples extracted from commits in which the code change also includes either a change to at least one return statement or the return type of the method. Furthermore, we discard examples involving methods that are longer than 30 lines, which is the 90th percentile for method lengths in the original primary dataset that we collect. Since it is unlikely that a `@return` comments can capture the essence of extremely long methods, we eliminate such cases. Moreover, most coding standards discourage such long methods, suggesting that these methods could be poorly written and possibly even poorly maintained. Additionally, the 90th percentile for the number of associated code tokens is approximately 40, and in order to reduce the number of cases in which there are substantial code changes that may be unrelated to the change in the comment, we remove such examples from our dataset. We also eliminate examples extracted from diffs involving more than 500 lines for a similar reason. To add, we disregard examples in which there are changes involving verb phrases in the comment as the code changes could be related to these phrases rather than the NP. We impose constraints to limit commits that involve insubstantial changes to the code or comment such as re-formatting, typo fixes, and simple rephrasing.

## C  Annotation Examples

We illustrate our annotation procedure in Figure 5. We consider only code tokens that are automatically labeled as associated and re-label any of these that we find to be irrelevant to the specified NP as not associated.

## D  Sample Output

We provide sample output from the binary classifier in Figure 6. In all examples, the bolded code tokens denote the tokens that the model predicts to be associated with the underlined NP in the comment and the highlighted ones indicate the true code tokens that the NP is associated with, as

```
/∗ @return a String representing the current active version of
    the ConfigStore.∗/
@Override
public String getCurrentVersion() {
    try {
        return storeMetadata.getCurrentVersion();
        } catch (IOException e) {
            Path configStoreDir = new Path(new Path(this.
                physicalStoreRoot),
                CONFIG_STORE_NAME);
            throw new RuntimeException(String.format("
                Error while checking current version for
                configStoreDir: \"%s\"", configStoreDir), e);
        }
}
```

(a) Example from from the eclipse-egit project.

```
/∗ @return the modified content∗/
public byte[] getModifiedContent() {
  byte[] result = new byte[modifiedContent.length];
  System.arraycopy(modifiedContent, 0, result, 0,
      modifiedContent.length);
  return result;
}
```

(b) Example from the apache-incubator-gobblin project.

Figure 5: Annotation example with all the bolded code tokens being automatically labeled as associated with the underlined NP in the comment and the crossed out tokens being manually re-labeled as not associated.

determined by manual annotation. For the example shown in Figure 6a, the model's predictions matches the gold associations in the annotated test set. In Figure 6b, the classifier accurately classifies the code token that is truly associated with the NP; however, it incorrectly identifies `Math` and `0` to be associated with the NP. This could be because the word "value" is often correlated with mathematical operations and numerical values, and so it could have possibly appeared in the training data with code tokens such as `Math` and `0`.

```
/∗ @return the SaveStepExecutionRes∗/
private SaveStepExecutionRes handleSaveStepExecution(
    SaveStepExecutionReq request) {
    SaveStepExecutionRes response = null;
    try {
        StepExecution stepExecution =
            JobRepositoryRpcFactory.
            convertStepExecutionType(request.
            stepExecution);
        stepExecutionDao.saveStepExecution(stepExecution
            );
        response = new SaveStepExecutionRes(
            stepExecution.getId(), stepExecution.getVersion
            ());
    } catch (Exception e) {
        log.error("error handling command", e);
    }
    return response;
}
```

(a) Example from the spring-yarn project.

```
/∗ @return The "advance" value or 0 if there is no text.∗/
private int getTextWidth(TextLayout textLayout) {
    if (textLayout != null) {
        return (int)Math.ceil(
            textLayout.getAdvance());
    }
    return 0;
}
```

(b) Example from the apache-pivot project.

Figure 6: Sample output of the binary classifier. The model classifies the bolded code tokens as associated with the underlined NP in the comment. The manually labeled, gold code tokens that are associated with the NP are highlighted.